\documentclass[10pt,twocolumn,letterpaper]{article}

\usepackage{cvpr}              % To produce the CAMERA-READY version
\definecolor{cvprblue}{rgb}{0.21,0.49,0.74}
\usepackage[pagebackref,breaklinks,colorlinks,allcolors=cvprblue]{hyperref}
\usepackage[T1]{fontenc}    % use 8-bit T1 fonts
\usepackage{booktabs}       % professional-quality tables
\usepackage{amsfonts}       % blackboard math symbols
\usepackage{nicefrac}       % compact symbols for 1/2, etc.
\usepackage{microtype}      % microtypography
\usepackage{xcolor}         % colors
\usepackage{colortbl}
\usepackage{tabu}
\usepackage{subcaption}
\usepackage{tikz}
\usepackage{wrapfig}
\usepackage{multirow}
\usepackage{caption}
\usepackage{multicol}
\usepackage{makecell}

\usepackage{pifont}

\newcommand{\greenmark}{\color{ForestGreen}\ding{51}}
\newcommand{\redxmark}{\color{red}\ding{55}}

\newcommand*\colourmark[1]{%
  \expandafter\newcommand\csname #1check\endcsname{\textcolor{#1}{\ding{51}}}%
}
\def\paperID{51} % *** Enter the Paper ID here
\def\confName{CVPR}
\def\confYear{2025}

\title{TimeLogic: A Temporal Logic Benchmark for Video QA}

\author{Sirnam Swetha\\
University of Central Florida\\
{\tt\small swetha.sirnam@ucf.edu}
\and
Hilde Kuehne\\
University of Tuebingen\\
{\tt\small h.kuehne@uni-tuebingen.de}
\and 
Mubarak Shah\\
University of Central Florida\\
{\tt\small shah@crcv.ucf.edu}
}

\begin{document}
\maketitle
\begin{abstract}
% \vspace{-0.15em}
Temporal logical understanding, a core facet of human cognition, plays a pivotal role in capturing complex sequential events and their temporal relationships within videos.
This capability is particularly crucial in tasks like Video Question Answering (VideoQA), where the goal is to process visual data over time together with textual data to provide coherent answers. 
However, current VideoQA benchmarks devote little focus to evaluating this critical skill due to the challenge of annotating temporal logic. %\hkc{there is a logical break here! is temporal reasoning a challenge or the evaluation of it?}
Despite the advancement of vision-language models, assessing their temporal logical reasoning powers remains a challenge, primarily due to the lack question-answer pairs that demand formal, complex temporal reasoning. 
To bridge this gap, we introduce the TimeLogic QA (TLQA) framework to automatically generate the QA pairs, specifically designed to evaluate the temporal logical understanding in VideoQA.
% To bridge this gap, we introduce a framework to automatically generate the QA pairs;the proposed Time Logic Dataset (TLD) Framework, is specifically designed to assess the temporal logical understanding capability in VideoQA.
To this end, TLQA leverages temporal annotations from existing video datasets together with temporal operators derived from \textit{logic theory} to construct questions that test understanding of event sequences and their temporal relationships. % within videos. 
As such, TLQA framework is generic and scalable, capable of leveraging both, existing video action datasets with temporal action segmentation annotations, or video datasets with temporal scene graph annotations, to automatically generate temporal logical questions.  %any existing video dataset with 076temporal annotations, including dense scene graph annotation traditional temporal segmentation datasets
% We assess the proposed setup by evaluating the temporal reasoning performance of current state-of-the-art VideoQA on 14 categories of temporal logic with varying temporal complexity, using data from four datasets: STAR, Breakfast, AGQA, and CrossTask.
%We assess the proposed setup by evaluating the state-of-the-art VideoQA models' temporal reasoning performance on 16 categories of temporal logic with varying temporal complexity for two question types, using data from four datasets: STAR, Breakfast, AGQA, and CrossTask. Furthermore, we generate two dataset variants – small and large – containing 1k and 5k question-answer pairs for each question type, respectively, for each category, resulting in 32k and 160k total pairs per dataset.
%moved at the end - We undertake a comprehensive evaluation ...
%Namely, 
We leverage 4 datasets, STAR, Breakfast, AGQA, and CrossTask, and generate two VideoQA dataset variants – small (TLQA-S) and large (TLQA-L) – containing 2k and 10k question-answer pairs for each category, resulting in 32k and 160k total pairs per dataset.
We undertake a comprehensive evaluation of leading-edge VideoQA models, employing the TLQA to benchmark their temporal logical understanding capabilities.
We assess the VideoQA model's temporal reasoning performance on \textbf{16} categories of temporal logic with varying temporal complexity.
% We assess the proposed setup by evaluating the VideoQA model's temporal reasoning performance on \textbf{16} categories of temporal logic with varying temporal complexity. %, demonstrating weak temporal understanding. % and demonstrate limitations in models temporal understanding. 
% Our evaluation focuses on model's ability to process and reason over the temporal information, we benchmark on various VideoQA models and demonstrate that current models have weak temporal understanding.
\vspace{-0.5cm}
\end{abstract}    
\section{Introduction}
\label{sec:intro}
% \vspace{-0.5cm}
At the intersection of vision and language, VideoQA poses a unique challenge: it requires models to accurately capture visual content over time and integrate textual information to generate coherent and precise answers. %it demands that a model not only accurately captures visual content over time, but also the integration of textual information to generate coherent and precise answers. 
This is especially challenging in the realm of temporal logical reasoning, a fundamental human skill that allows deducing information, resolving ambiguities, and answering questions based on a %to deduce information, resolve ambiguities, and answering questions based on a %combination of inputs and 
sequence of events in a video.
% previous knowledge. \hkc{you don't do action prediction or anticipation, right?}
%This task goes beyond merely recognizing objects and their spatial relationships; it necessitates a nuanced grasp of how these relationships evolve through time, influenced by a sequence of actions and events. 
Temporal logical reasoning goes beyond merely recognizing objects and their spatial relationships; it necessitates a nuanced understanding of how these relationships evolve over time, influenced by a sequence of actions and events. 
For instance, to answer a question like ``Which action always occurs before person opening a cabinet, which in turn always occurs before person holding the clothes?'' %"What does the person do while talking on the phone but before opening the door?", 
the model must parse and reason about the temporal order and logical connections between the actions depicted in the video.

However, annotating temporal logic is challenging for humans, therefore existing VideoQA benchmarks ~\cite{wu2021star_situated_reasoning, Yi2020CLEVRER, xiao2021next, GrundeMcLaughlin2021AGQA, yu2023anetqa} devote only a small fraction of their questions to this task, primarily focusing on simple logical constructs such as `and', `or', `not', `before', and `after'. As a result, these benchmarks provide a limited assessment of complex temporal understanding within videos. While current VideoQA benchmarks are impressive in scope, they predominantly cater to questions revolving around objects, fine-grained attributes, and simple ordering-based logical constructs, often failing to encapsulate full spectrum of complexity inherent in temporal reasoning of video data as shown in Table~\ref{tab:relworks_data}.

\begin{table*}
% \vspace{-1cm}
  \centering
  \resizebox{0.95\textwidth}{!}{
  \begin{tabu}{lccccccccccc}
    \toprule
       Temporal & Complexity & Syntax & CLEVRER & MarioQA & TGIF-QA & ActivityNet-QA & STAR &  AGQA & NExT-QA &AnetQA & TLQA  \\
       
       Operators & Level &  & \cite{Yi2020CLEVRER} & \cite{mun2017marioQA} & \cite{jang2017tgif} & \cite{yu2019activityqa} & \cite{GrundeMcLaughlin2021AGQA} & \cite{GrundeMcLaughlin2021AGQA}  & \cite{xiao2021next} & \cite{yu2023anetqa} & (ours) \\      
    \midrule
    Real-world data &  &  & \redxmark & \redxmark & \greenmark  & \greenmark & \greenmark & \greenmark &  \greenmark &  \greenmark & {\bf \greenmark}  \\   
    \midrule
    Eventual & 1 & E(X) &  \greenmark & \greenmark & \greenmark & \greenmark & \greenmark  & \greenmark & \greenmark & \greenmark  & \greenmark \\
    Always & 2 & G(X) &      \redxmark      &     \redxmark      &     \redxmark &     \redxmark       &     \redxmark       &     \redxmark      &     \redxmark      &     \redxmark      &  \greenmark \\
    Until & 3 & X $\cup$ Y  & \redxmark &  \redxmark&\redxmark  & \redxmark &     \redxmark   & \redxmark   & \redxmark &     \redxmark       & \greenmark  \\      
    Since & 3 & X S Y  &  \redxmark& \redxmark & \redxmark & \redxmark &     \redxmark    & \redxmark  & \redxmark &     \redxmark      & \greenmark  \\    
    Disjoint & 3 & X D Y  & \redxmark & \redxmark & \redxmark & \redxmark & \redxmark  & \redxmark & \redxmark  & \redxmark & \greenmark \\    
    Implies & 3 & X $\rightarrow$ Y  & \redxmark & \redxmark &\redxmark  & \redxmark &\redxmark  &  \redxmark      &    \redxmark     &     \redxmark     & \greenmark \\ 
    Co-Occur & 3 & X $\wedge$ Y & \redxmark & \redxmark & \redxmark & \redxmark &  \redxmark & \redxmark &\redxmark &  \redxmark & \greenmark \\ 
    %Overlap &  &  &  &  &  &  &   &  &  &   \\      
    Next/Before & 3 & X > Y/ X < Y & \greenmark & \redxmark  & \greenmark & \greenmark & \greenmark  &  \greenmark & \greenmark & \greenmark  & \greenmark  \\      
    Immediately Next/Before & 4  & X IA Y/X IB Y & \greenmark & \redxmark & \redxmark &    \redxmark &   \redxmark      &   \redxmark      &   \greenmark*      &    \redxmark     &  \greenmark \\      
    Always (Next/Before) & 4 &  X AA Y/A AB Y & \redxmark & \redxmark &  \redxmark &  \redxmark &   \redxmark  &   \redxmark      &    \redxmark       &     \redxmark    &    \greenmark  \\  
    % Always Past & 4 &  &  &  &  & no & no  & no & no  &  yes \\  
    % Always Past &  &  &  &  &  &  &   &  &   &  \\  
    
    Always Co-Occur & 4 & X AC Y & \redxmark & \redxmark & \redxmark  &\redxmark & \redxmark &   \redxmark   & \redxmark & \redxmark &  \greenmark \\ 
    \midrule
    \multicolumn{11}{c}{\textbf{Compounding}} \\
    \midrule 
    Strict Ordering X,Y,Z & 5 & X AB Y AB Z & \redxmark  & \redxmark &   \redxmark  &   \redxmark   &   \redxmark     &  \redxmark     &     \redxmark    &    \redxmark      & \greenmark \\  
    Loose Ordering X,Y,Z & 5 & X < Y < Z & \redxmark & \redxmark &  \redxmark &   \redxmark   &   \redxmark   &    \redxmark     &  \redxmark     &     \redxmark        & \greenmark \\  
    X always before Y,Z & 5 & X < (Y, Z)  & \redxmark &   \redxmark  & \redxmark &     \redxmark   &   \redxmark  &  \redxmark    &  \redxmark    &     \redxmark   & \greenmark \\  
    \bottomrule
  \end{tabu}
  }
  \vspace{-0.2cm}
  \caption{Comparison of temporal complexity of current VideoQA datasets compared to our proposed TLQA dataset including respective complexity levels. * indicates very few questions.  %Here, - represents No.  
  }
\label{tab:relworks_data}
\vspace{-0.2in}
\end{table*}

Recognizing these challenges,% our research presents 
we introduce a new framework the Time Logic QA (TLQA) framework 
as well as a respective benchmark called Time Logic QA (TLQA) benchmark, aimed explicitly at evaluating the temporal understanding ability of VideoQA models and systems. % a meticulously curated collection of VQA tasks that are designed to specifically probe models for their temporal reasoning skills.  \hkc{more system description here:} 
Our framework generates QA pairs for each of the 16 temporal logic categories shown in Table~\ref{tab:relworks_data} with pre-defined template questions. The temporal categories in TLQA benchmark can be organized into 5 complexity levels, requiring multiple inference steps with  level 1 being the least complex and level 5 being the most complex.
%Our framework generates two types of QA pairs, (i) Boolean and (ii) Multiple-Choice QA for each of the 16 temporal logic categories shown in Table~\ref{tab:relworks_data} with pre-defined template questions. 
%This enables deterministic evaluation of models, unlike open-ended QA that utilizes external judges like GPT to compute accuracy~\cite{}, which can be inconsistent. % and biased.
The proposed TLQA framework is generic and scalable to any existing video dataset with temporal annotations, including dense scene graph annotations utilized in VideoQA datasets such as STAR~\cite{wu2021star_situated_reasoning}, AGQA~\cite{GrundeMcLaughlin2021AGQA}, AnetQA~\cite{yu2023anetqa} or traditional temporal segmentation datasets like Breakfast~\cite{Kuehne12bf} and CrossTask~\cite{Zhukov2019crosstask}. The proposed framework can transform any video dataset with temporal annotations into temporal logic QA pairs.

The TLQA benchmark embodies a structured suite of questions, each crafted to encapsulate temporal operators that are rooted in logic theory. These operators are tools through which the dataset examines a model's capacity to decipher the order, concurrency, and causality of events as depicted in video content. By integrating a systematic analysis of temporal logic within video content, TLQA aspires to fill a critical gap in current benchmarks, offering a more challenging and comprehensive testbed that mirrors the temporal complexity and logical depth of real-world events. TLQA comprises of 16 temporal logic categories and 32k/160k QA pairs, including 16k/80k Boolean questions and 16k/80k multiple-choice questions per dataset, combining for four datasets leads to cumulative of 128k/640k QA pairs, providing a comprehensive and challenging benchmark for evaluating temporal logical understanding in video analysis.

%This paper
We undertake a comprehensive evaluation of leading-edge VideoQA models, employing the TLQA to benchmark their temporal logical understanding capabilities.  We perform Zero-Shot evaluation on current VideoQA models across various architectures with differing temporal capacities, including models specialized for QA tasks, expert models, and caption-based QA. 
Our results indicate that while current models perform relatively well on the multiple-choice task, they struggle significantly with boolean questions. Additionally, models tend to be dataset agnostic for boolean questions, and there are variations in the percentage of 'yes' responses across temporal categories.
% Our results show that current models, while performing  .... on the multiple choice task, struggle on binary ... \hkc{Anything from eval... MC better than binary,  models are dataset agnostic for binary? , different yes ratio}
% We dissect the results of this evaluation, revealing insights into where current approaches excel and where they fall short. 
By highlighting these findings, we underscore the necessity for advancements in the modeling of temporal understanding. 
% It is our contention that the future of VQA, and by extension video comprehension, lies in the development of models that can navigate the temporal dimension with the same agility that humans do. 

% Our contribution through the TLQA dataset is two-fold: (i) TLQA aims to provide a more rigorous testbed for evaluating the temporal logical understanding capabilities of VideoQA models, with a particular focus on their ability to perform complex inference and multi-step temporal understanding over video content. 
% (ii) The proposed framework to automatically generate QA pairs utilizing existing Video datasets without needing specialized annotations, our framework is generic and scalable to diverse datasets.
% (iii) By systematically analyzing and incorporating temporal logic into the generation of QA pairs, TLQA seeks to address the limitations of existing datasets and offer a comprehensive benchmark that reflects the diversity and complexity of real-world events. Through this endeavor, we aspire to not only advance the field of VideoQA but also contribute to the broader goal of developing systems with more profound logical understanding and comprehension abilities.

\noindent Our key contributions are: (i) We propose the TLQA framework to automatically build temporal logical QA pairs at scale from existing video datasets and their annotations, (ii) we propose the TLQA benchmark, a comprehensive set of temporal QA pairs that enables evaluating temporal logical reasoning abilities across multiple levels of complexity and datasets, and (iii) we  %show an extensive evaluation and analysis of
evaluate current state-of-the-art models for temporal Video QA and provide extensive evaluation.

\vspace{-0.15cm}
\section{Related Works}
\noindent \textbf{VideoQA Benchmarks} VideoQA has gained significant attention in recent years, with several benchmarks proposed to evaluate model capabilities~\cite{Yi2020CLEVRER, mun2017marioQA, jang2017tgif, yu2019activityqa, GrundeMcLaughlin2021AGQA, xiao2021next, yu2023anetqa, wu2021star_situated_reasoning}. Those benchmarks cover a wide range of questions about objects and actions involving fine-grained attributes, simple causal relationships, as well as spatio-temporal reasoning. However, while temporal processing is one of the main features that distinguishes video from image-based QA, VideoQA benchmarks consider only a fraction of the questions to evaluating temporal logical reasoning abilities, e.g. STAR features 25\% sequencing questions, NextQA reports 29\% of temporal QA pairs, and ANetQA reports 20.05\% sequence and 0.05\% duration questions, also noting that those types of questions have the lowest performance of all questions from this type. Additionally, as shown in Table~\ref{tab:relworks_data}, those questions primarily focus on simple logical constructs like 'and', 'or', 'not', 'before', and 'after'. 
In terms of automatic generation of temporal QA pairs for video data, various approaches have been leveraged so far. CLEVRER~\cite{Yi2020CLEVRER} generates complex reasoning questions on synthetic data, also including temporal prediction. STAR~\cite{wu2021star_situated_reasoning} also uses automatize scripts to generate QA pairs from situation hypergraphs, including questions about sequences. In a similar context, AGQA~\cite{GrundeMcLaughlin2021AGQA} proposes handcrafted programs that operate over the annotated scene graphs to also generate questions and answers automatically. %But 
However, even those scripts usually only cover a small fraction of all possible temporal logic operators. Compared to those work, we systematically combine temporal logic operators with existing annotation for temporal logic question generation, providing a more exhaustive coverage of possible temporal correlations than previous works. 

\begin{table*}
  \centering
  \resizebox{0.95\textwidth}{!}{
  \begin{tabu}{lccc}
    \toprule
    % & & \multicolumn{4}{c}{VQAv2 Accuracy}                   \\
    % \cmidrule(r){3-6}
       \textbf{Temporal} & \textbf{Syntax} & \textbf{Description} & \textbf{Definition}  \\
       \textbf{Operator} &  &  &   \\      
    \midrule
    Eventual & E(X)  & Eventually X will be true at some point & M, t $\models$ E(X)\ if\ $\exists$ t' > t : M, t' $\models$ X \\
    Always & G(X)  & X is always true at all points in time & M, t $\models$ G(X) $\iff$ $\forall$ t' $\geq$ t, M, t' $\models$ X \\
    Until & X U Y & X is true at every moment before Y becomes true %A occurs until B takes place. A is true at every moment before B becomes true 
    & M, t $\models$ X\ U\ Y $\iff$ $\exists$ t' > t : M, t' $\models$ Y\  and\  $\forall$ t'' (t < t'' < t') M, t'' $\models$ X  \\      
    Since  & X S Y & X has been true since a time where Y was true & M, t $\models$ X\ S\ Y $\iff$ $\exists$ t' < t : M, t' $\models$ Y\  and\  $\forall$ t'' (t' < t'' < t) M, t'' $\models$ X  \\  
    Disjoint & X D Y & Both X and Y cannot be true at the same time & M, t $\models$ X\ D\ Y\ if\ M, t $\models$ $\neg$ X\ or\ M, t $\models$ $\neg$  Y \\ 
    Implies  & X $\rightarrow$ Y  & If X is true, then Y must also be true &  M, t $\models$ X $\rightarrow$ Y\ if\ M, t $\models$ $\neg$ X\ or\ M, t $\models$ Y \\ 
    %Overlap &  &  &  &  &  &  &   &  &   \\      
    Before  & X < Y & X occurs before Y	 & M, t $\models$ X < Y $\iff$ $\exists$ t' (t < t') : M, t $\models$ X\ and\ M, t' $\models$ Y\ and\  $\neg$ $\exists$ t'' (t < t'' < t') M, t'' $\models$ Y   \\  
    After & X > Y & X occurs after Y & M, t $\models$ X > Y $\iff$ $\exists$ t' (t' < t) : M, t' $\models$ Y and M, t $\models$ X and $\neg$ $\exists$ t'' (t' < t'' < t) M, t'' $\models$ X \\
    Immediate Before  & X IB Y &  X is true immediately before Y & M, t $\models$ X\ IB\ Y\ if\ M, t-1 $\models$ X\ and\ M, t $\models$ Y   \\      
    Immediate Next  & X IA Y &  X is true immediately after Y & M, t $\models$ X\ IA\ Y\ if\ M, t+1 $\models$ X\ and\ M, t $\models$ Y   \\      
    Always Before  & X AB Y & Whenever Y is true, X was always true before  & M, t $\models$ X\ AB\ Y\ if\ $\forall$ t' (M, t' $\models$ Y)\ implies\ $\forall$ t'' < t', M, t'' $\models$ X  \\  
    Always Next  & X AA Y &  Once Y is true, X is always true afterwards & M, t $\models$ X\ AA\ Y\ if\ $\forall$ t' (M, t' $\models$ Y)\ implies\ $\forall$\ t'' > t', M, t'' $\models$ X  \\  
    % Always Past  & H(X) &  X has always been true in the past &  M, t $\models$ H(X)\ if\ $\forall$ t' < t, M, t' $\models$ X \\  
    % Always Past &  &  &  &  &  &  &   &  &   \\  
    Co-Occur  & X $\wedge$ Y & X and Y occur simultaneously &  M, t $\models$ X $\wedge$ Y $\iff$ M, t $\models$ X\ and\ M, t $\models$ Y \\ 
    Always Co-Occur  & X AC Y & X and Y always occur together & M, t $\models$ X\ AC\ Y\ if \ $\forall$ t', M, t' $\models$ X $\iff$ M, t' $\models$ Y  \\ 
    \midrule
     \multicolumn{4}{c}{\textbf{Compounding}} \\
    \midrule 
    Strict Ordering X,Y,Z  & X AB Y AB Z   & \makecell{X always occurs before Y, \\which in-turn always occurs before Z} &  M, t $\models$ X\ AB\ Y\ AB\ Z\ if\ $\forall$ t' (M, t' $\models$ X)\ implies\ ($\forall$ t'' < t', M, t'' $\models$ Y\ and\ $\exists$ t''' < t'', M, t''' $\models$ Z) \\  
    Loose Ordering X,Y,Z  & X < Y < Z &  X occurs before Y and Y occurs before Z &  M, t $\models$ X\ <\ Y\ <\ Z\ if\ ($\forall$ t' (M, t' $\models$ Z)\ implies\ $\forall$ t'' < t', M, t'' $\models$ X) and\ ($\forall$ t' (M, t' $\models$ Y)\ implies\ $\forall$ t'' < t', M, t'' $\models$ Z)  \\  
    X always before Y,Z  & X < (Y, Z) & X always occurs before Y, Z  &  M, t $\models$ X\ <\ (Y , Z)\ if\ ($\forall$ t' (M, t' $\models$ Z)\ implies\ $\forall$ t'' < t', M, t'' $\models$ X) and\ ($\forall$ t' (M, t' $\models$ Y)\ implies\ $\forall$ t'' < t', M, t'' $\models$ X) \\  
    \midrule
    \bottomrule
  \end{tabu}
  }
  \vspace{-0.1cm}
  \caption{Overview of temporal operators syntax. M indicates the temporal model, t indicates time.}
  \label{tab:syntax}
  \vspace{-0.2cm}
\end{table*}

\noindent \textbf{VideoQA models}
While early works on VideoQA~\cite{Zhu2017UncoveringTC, jang2017tgif, Yi2020CLEVRER} focused on building elaborated reasoning architectures, this trend changed with the increased availability and performance of language models. As a result, current architectures explore different ways to combine video input with text based processing.
SeViLA~\cite{yu2024self} utilizes BLIP-2 to create special localizer and reasoning modules in order to build a VideoQA system. LLoVI~\cite{zhang2023simple} leverages a video captioner and a text LLM is supplied with the captions to answer questions. These methods are specialized for VideoQA tasks and are not truly general VideoLLMs.
Video LLaMA~\cite{zhang2023video} was one of the first methods to extend BLIP-2 to build a video LLM using dual video and audio Q-Formers and trained it on image-caption \& video-caption data. Video-LLaVA~\cite{lin2023video} achieves strong multi-modal performance by utilizing the LanguageBind vision model unifying representations of images and videos, projected into a shared language feature space.  Video-ChatGPT~\cite{maaz2023video} built a high-quality Video instruction tuning dataset by utilizing strong vision foundation models to extract semantic information from the videos and using an LLM to generate questions and answers from this corpus. VideoChat~\cite{li2023videochat} utilizes foundation models to generate video captions, tags etc and utilizes them as additional input along with video features. LLaMA-VID~\cite{li2023llama} uses a context attention module to reduce each frame of the video down to 2 tokens, permitting it to use a large number of input frames per video. Chat-UniVI~\cite{jin2023chat} utilizes token merging methods to achieve the same goal. ImageGrid-VLM~\cite{kim2024igvlm} and PLLaVA~\cite{xu2024pllava} take a different approach by passing videos as image grids or filmstrips to pre-trained Image VLMs. %LLaVA-NeXT~\cite{zhang2024llavanextvideo} demonstrates that multi-resolution trained Image VLMs can also understand Video with no additional training. LLaVA-Hound~\cite{zhang2024direct} carries out DPO training on a LLaVA-NeXT model with high quality video captions generated by GPT4-V.

%For benchmarking, we do zero-shot evaluations with VideoLLaVA~\cite{lin2023videollava}, VideoChatGPT~\cite{Maaz2023VideoChatGPT}, ChatUnivi~\cite{jin2023chatunivi}, SeViLA~\cite{yu2023sevila}, IGVLM~\cite{kim2024igvlm}, LLoVI~\cite{zhang2023llovi}. Note for all LLaVA-based~\cite{liu2023llava} models we use the 7b backbone and SeViLA employs BLIP-2~\cite{li2023blip2} with FlanT5XL.

\vspace{-0.1cm}
\noindent \textbf{Evaluation Protocols} Finally, evaluating the performance of VideoQA models is a critical aspect of benchmarking. Existing benchmarks include the following question types: boolean, i.a.~\cite{xiao2021next, yu2023anetqa}, multiple-choice~\cite{wu2021star_situated_reasoning, Yi2020CLEVRER, xiao2021next} and open-ended questions~\cite{xiao2021next, yu2023anetqa} which are evaluated by matching the generated answer to the ground truth. Since, this evaluation is straightforward for boolean and multiple-choice as they can be directly matched with ground-truth, we chose to generate boolean and multiple-choice QA pairs for benchmarking. For open-ended answers, they can be incorrectly classified wrong if not matched exactly to the ground-truth text. To address this, external judges like ChatGPT is employed to compute accuracy, but it can be inconsistent and susceptible to biases. Also, different models support different form of QA. While some models resp QA tasks require fine-tuning for a specific target vocabulary, models based on LLMs~\cite{lin2023videollava,Maaz2023VideoChatGPT,jin2023chatunivi}, can handle any form of open ended QA.
\vspace{-0.1cm}
\section{TLQA Framework: Time-Logic QA}
\vspace{-0.15cm}
\begin{figure*}
    \centering
    \includegraphics[width=1.0\textwidth]{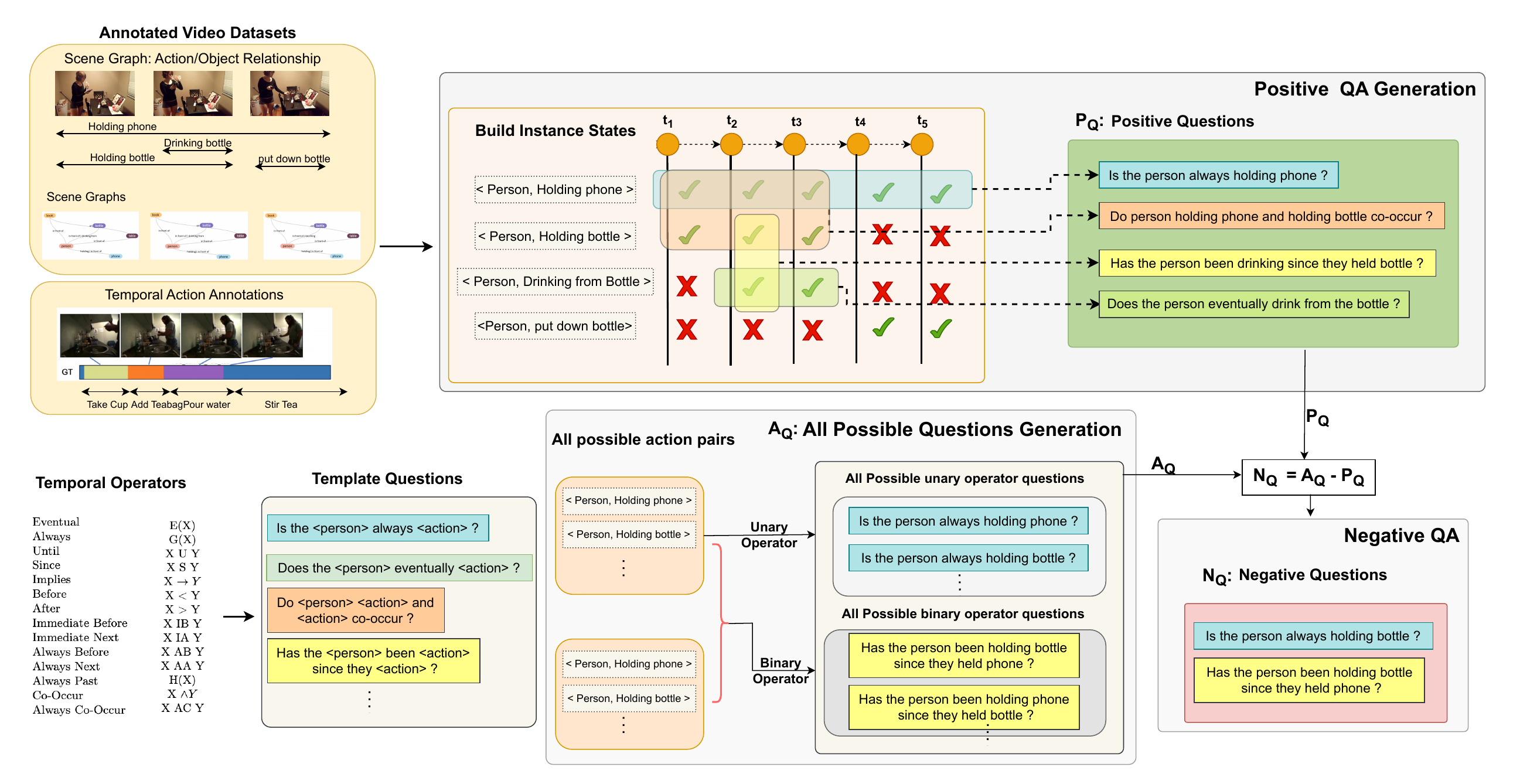}
      \caption{Framework Overview. Given existing video datasets with either dense scene graph annotations or temporal action annotations as illustrated, our framework will automatically generate QA pairs for temporal logic with varying complexity. First, we build instance states representing the overall action at each time step ($t_i$) throughout the video as shown. For each temporal category as shown, we generate all positive questions ($P_Q$) valid for the video satisfying the temporal logic definition. Then we generate all possible questions ($A_Q$) by taking all possible action for the video/dataset. The negative questions are the sampled from  $A_Q$-$P_Q$ .}
  \label{fig:pipeline}
  
\end{figure*}

The TLQA benchmark is generated automatically by leveraging existing VideoQA datasets with either dense temporal scene graph annotations~\cite{wu2021star_situated_reasoning, GrundeMcLaughlin2021AGQA} or temporal action localization/segmentation datasets~\cite{Kuehne12bf, Zhukov2019crosstask} with temporal annotations for actions. Utilizing these annotations as a foundation, we generate questions related to actions and their ordering. Given the temporal logic operators, we define template questions for each of the temporal category as shown in Table~\ref{tab:templates_format}.
We offer 2 versions of the dataset for small/large scale evaluation, encompassing 16 categories spanning across complexity levels 1 to 5, totaling 32k/160k QA pairs per dataset. Accurately answering these questions requires complex temporal logical understanding and multi-step inference.

The automatic creation of the TLQA benchmark involves: (1) defining the temporal logic categories; (2) creating template questions for each category; (3) building instance states for any video dataset with temporal annotations; (4) automatically generating positive QA pairs for each category; and (5) automatically generating negatives for both boolean and multiple-choice questions based on the positive QA pairs.

% The TLQA dataset is generated by automatically leveraging existing Video QA datasets which have either dense scene graph annotations for object, action and attribute relationships (~\cite{wu2021star_situated_reasoning, GrundeMcLaughlin2021AGQA, yu2023anetqa}) or temporal action localization datasets ~\cite{Kuehne12bf, Zhukov2019crosstask} which have temporal annotations for actions in videos. Utilizing these annotations as a foundation, we generate questions related action, object relationship or solely focus on actions. We provide 3 versions of the dataset for small-scale, medium-scale and large-scale evaluation. It encompasses 16 categories, which are distributed across a complexity range from 1 to 5. Answering the questions accurately requires complex temporal logical understanding demanding multi-step inference. 

\subsection{Formal Temporal Logic Definition}
\begin{figure}
\vspace{-1.8em}
% \hspace{-1em}
    \centering
    \includegraphics[width=0.4\textwidth]{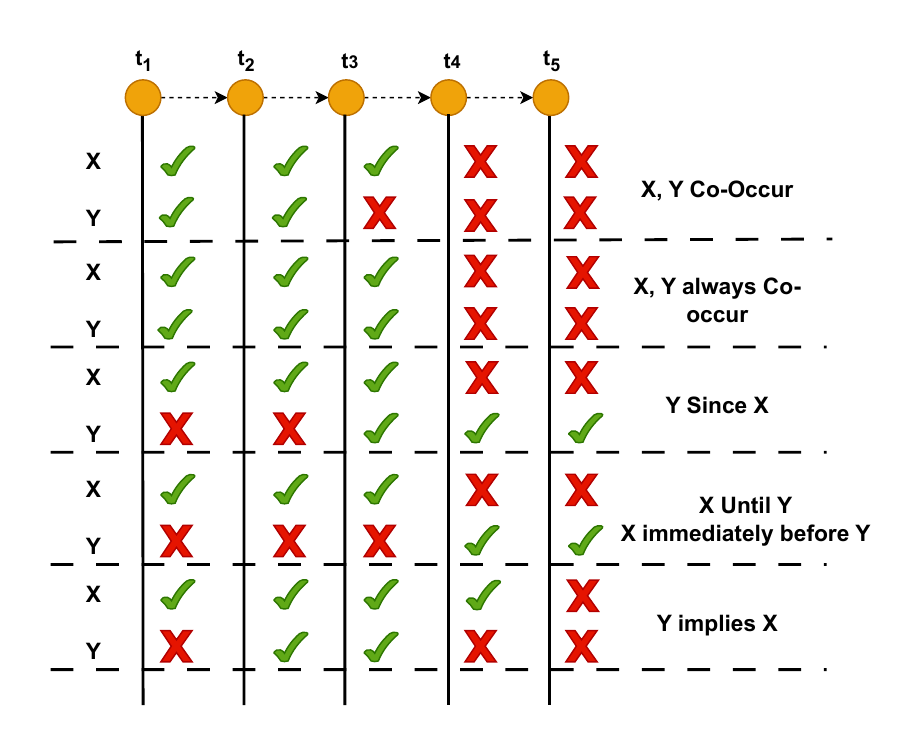}
    \vspace{-1.2em}
      \caption{Temporal Intervals for two actions X, Y. $t_i$: time step.}
      \vspace{-1.5em}
  \label{fig:intervals}
\end{figure}

Temporal logic is an extension of Description Logic for formal representations of temporal relationships. In Table.~\ref{tab:syntax} we present the temporal operators along with their syntax and formal definitions for each of the 16 categories. For simplicity, we show it pictorially in Figure.~\ref{fig:intervals} illustrating their application for two actions. Note that $X$ $U$ $Y$ is the same as X immediately occuring before Y as shown in Figure.~\ref{fig:intervals}. We take the most common occurring temporal order variants observed in real-world situations.  
% For each syntax in the Table.~\ref{tab:syntax}, we define template questions as shown in Table.~\ref{tab:templates_format} to form our QA template set for both boolean and multiple-choice questions. These templates form the backbone of our QA template set guiding the automatic generation of questions.
\subsection{Template Questions with Temporal Operators}
As shown in  Table.~\ref{tab:syntax}, the temporal categories involve unary and binary temporal operators.
% We first identify temporal operators that are more relevant for real-world action videos and define templates for each of them with both unary and binary temporal operator.
To further add more complexity beyond unary and binary temporal operators, we meaningfully compound them and select the most relevant combination that commonly occur in real-world videos. Table~\ref{tab:templates_format} shows the template questions for each category for boolean question type. Please refer to supplementary Section 4 for multiple-choice template questions.
In order to generate the questions with proper english syntax structure, we define tenses for all actions to accurately represent past, present, future, continuous, and perfect tenses. For e.g.: `Has the <person> been <action0 in present perfect continuous tense> since they <action1 in simple past tense> ?'. %In Table.~\ref{tab:templates_format}, we show the templates for all the categories. % for both boolean and choice based questions.

\paragraph{Question Types} For deterministic evaluation of VideoQA, we categorize questions into two types: Boolean and Multiple-Choice. Boolean questions require the model to respond with a simple `yes' or `no'. In contrast, Multiple-Choice type questions present the model with multiple options, from which it must select the one that correctly answers the question. The following sections delve into the methodologies for multiple-choice selection and question sampling, ensuring a systematic evaluation framework.

% action and attributes
% bool and what-choice for deterministic evaluation

\begin{table*}\setlength{\tabcolsep}{5pt}
  \centering
  \resizebox{0.85\textwidth}{!}{
  \begin{tabular}{lll}
    \toprule
      \textbf{ Temporal Operators} & \textbf{Complexity} &  \textbf{Template Questions }  \\
       % Operators &  &   \\      
    \midrule
    Eventual & 1 & Does the <person> eventually <action> ? \\%,  What does the <person> do in this video ? \\
    Always & 2 & Is the <person> always <action> ? \\ %,  What action does the <person> do throughout the video ? \\ % $<{action}\_0>$
    Until & 3 & Did the <person> <action> until <action> ?  \\
    Since & 3 & Has the <person> been <action> since they <action> ?  \\
    Disjoint & 3 & Is it true that <person> <action> does not overlap with <action> ?  \\
    Implies & 3 & Does the <person> <action> imply <action> ?    \\
    %Overlap &  & &   &  &   &   & \\      
    Before &  3& Did the <person> <action> before <action> ? \\
    Next & 3 & Did the <person> <action> after <action> ?  \\
    Co-Occur & 3 & Do <person> <action> and <action> co-occur ? \\
    % Immediate Before &  4 &  \\
    Immediate Next & 4  & Did the <person> <action> immediately after <action> ?   \\
    Always Before & 4 & Did the <person> <action> always before <action> ?  \\
    Always Next & 4 & Did the <person> <action> always after <action> ?   \\
    Always Co-Occur & 4 &  Is it true that <person> <action> always co-occur with <action> ? \\
    \midrule
     \multicolumn{3}{l}
     {\textbf{Compounding}} \\
    \midrule 
    Strict A,B,C & 5 &  \makecell{Is it true that <person> <action>  always occurs before <person> <action>, \\ which in turn always occurs before <person> <action> ?}\\
    
    Loose A,B,C &  5&  \makecell{Is it true that <person> <action> occurs before <person> <action>, \\ which in turn occurs before <person> <action> ?}\\
    A always before B,C & 5 &  \makecell{Is it true that <person> <action> always occurs before <person> <action> \\ and <person> <action> ?} \\
    \bottomrule
  \end{tabular}
  }
  \caption{Illustration of template questions for each category of the Temporal Logic Dataset.}
  \label{tab:templates_format}
  \vspace{-0.2in}
\end{table*}

\subsection{Building Instance States: Re-purposing Existing Annotated Datasets}
Our framework is generic and scalable to multiple datasets, leveraging traditional video action datasets with temporal annotations such as Breakfast~\cite{Kuehne12bf}, CrossTask~\cite{Zhukov2019crosstask}; while also extending to prior VideoQA datasets which have dense annotations such as STAR~\cite{wu2021star_situated_reasoning}, AGQA~\cite{GrundeMcLaughlin2021AGQA}. %The annotations involve object attributes along with actions to generate more complex questions relating to both object and actions such as STAR~\cite{wu2021star_situated_reasoning}, AGQA~\cite{GrundeMcLaughlin2021AGQA}, ANetQA~\cite{yu2023anetqa}. 
We discuss the dataset details in Section~\ref{sec:data_details}.

Given the dense scene graph/temporal annotations, we build comprehensive instance states capturing object and action at each time step in the video, as shown in Figure~\ref{fig:pipeline}. By examining the instance states in the video, for example we can infer that the action `holding phone' occurs throughout the video as illustrated in Figure~\ref{fig:pipeline}. We now define the automated QA generation process given the instance states.

% \subsubsection{Define template questions with temporal operators}

% We first identify temporal operators that are more relevant for real-world action videos and define templates for each of them with both unary and binary temporal operator.
% To further add more complexity beyond unary and binary temporal operators, we do meaningful compounding of them and select the most relevant combination that commonly occurs in real-world videos.

% In order to generate the questions with proper english syntax structure, we define tenses for all the actions to represent past, present, future, continuous, perfect tenses accurately. For eg: `Has the <person> been <action0 in present perfect continous tense> since they <action1 in simple past tense> ?'. In Table.~\ref{tab:templates_format}, we show the templates for all the categories. % for both boolean and choice based questions.

\subsection{Automatic Positive QA Generation}

Given the template questions for a temporal category, we now present the fully automated positive QA generation process leveraging the instance states generated from video datasets with temporal annotations. 
For each video, we compute all unique actions occurring in it and represent each action with both `actor' and `action'. If the annotation for actor is missing (as in the case of Breakfast/CrossTask which only have action annotations), we use `person' to denote actor. We also store all unique actions in the dataset to represent the full action set.
Given a template question and all instance states in a video, we implement a dynamic programming solution to find all valid/positive questions for the temporal operators in the template question. We define scripts for each temporal category to recursively build the set of positive questions satisfying the temporal logic as per the definition in Table~\ref{tab:syntax}.
First, we generate all positive instance-level objectives at each time-step from the instance states, i.e., at time step t, we have all positive actions set. Then, we recursively build the set of positive actions satisfying the temporal operators in a given category by implementing the logic for each temporal operator. % To limit the number of questions, we sample questions per video
For example, to generate all positive questions for eventual category, at each time-step, we take the actions list and merge them recursively to get unique positive actions for the category. To transform them into positive questions, we replace the <actor> and <action> in the template question with the computed unique positive actions, and the answer is set to `yes' for boolean questions. For multiple-choice questions, the positive actions form the correct answer choice which we position randomly at different positions (a,b,c,d)  to have a uniform answer-choice distribution.
%We first generate all instance-level questions that are valid at each time-step ie., does the <actor> perform the <action> ? Then we recursively build the positive questions satisfying the temporal operators, we implement logic for each temporal operator
Similarly, for the always temporal category, to satisfy the constraint, the action \textit{must} occur in \textit{all} time steps. We implement this constraint while computing the unique positive actions for each video following the definition in Table~\ref{tab:syntax}. 

\begin{table*}
    \centering
    \resizebox{\textwidth}{!}{
    \begin{tabu}{cccccccccccccc}
    \toprule
    Temporal & & \multicolumn{3}{c}{STAR} & \multicolumn{3}{c}{Breakfast} & \multicolumn{3}{c}{AGQA} & \multicolumn{3}{c}{CrossTask} \\
    \cmidrule(lr){3-5} \cmidrule(lr){6-8}  \cmidrule(lr){9-11} \cmidrule(lr){12-14} %  \cmidrule(lr){12-14} 
    Operators & Complexity & SeViLA & IGVLM & LLoVI & SeViLA & IGVLM & LLoVI & SeViLA & IGVLM & LLoVI & SeViLA & IGVLM & LLoVI\\
    \midrule
    Eventual & 1 & 59.1 & 34.5 & 42.5 & 34.4 & 22 & 31.6 & 59.2 & 21 & 44.6 & 76.8 & 40 & 37.1 \\
    Always & 2 & 58.8 & 43.8 & 51.8 & - & - & - & 63.6 & 35.7 & 50.9 & - & - & - \\
    \midrule
    Until & 3 & 55.8 & 56.1 & 51.5 & 37.4 & 33.6 & 29.4 & 58.7 & 54.8 & 53.3 & 72 & 63.3 & 38.5 \\
    Since & 3 & 57.3 & 45.1 & 45.6 & - & - & - & 57.8 & 39.2 & 39.6 & - & - & - \\
    Implies & 3 & 67.7 & 68.6 & 62.2 & - & - & - & 67.5 & 61.3 & 54.6 & - & - & - \\
    Before & 3 & 55.1 & 58.6 & 49 & 44.6 & 36 & 36.6 & 60.5 & 56.8 & 52.6 & 72 & 63.5 & 50.2 \\
    Next & 3 & 50.2 & 42.9 & 33 & 38.8 & 30.5 & 31.9 & 63.5 & 43.2 & 48.2 & 78.5 & 61.6 & 50.3 \\
    Co-Occur & 3 & 61 & 43.2 & 57.1 & - & - & - & 57.1 & 33.7 & 49.7 & - & - & - \\
    Disjoint & 3 & 32.7 & 43.1 & 43 & 26.1 & 28.9 & 33.1 & 32.4 & 42.2 & 45.2 & 31 & 51.1 & 42.2 \\
    \midrule
    Immediate Next & 4 & 49.4 & 40.3 & 35 & 41 & 32.4 & 29.9 & 63.2 & 41.6 & 43.2 & 78.5  & 53  & 64.7 \\
    Always Before & 4 & 52.5 & 57.6 & 52.2 & 39.5 & 41.6 & 35 & 58.6 & 55.4 & 52.9 &  62.1& 62.5 & 42.3 \\
    Always Next & 4 & 51.8 & 49 & 47.9 & 42.8 & 34.6 & 34.2 & 59.8 & 44 & 48.7  & 62.6 & 52.7 & 50.3 \\
    Always Co-Occur & 4 & 69.3 & 68.1 & 69.8 & - & - & - & 68.7 & 67.2 & 72.2 & - & - & - \\
    \midrule
    Strict A,B,C & 5 & 48.4 & 60 & 60.5 & 48.1 & 40.7 & 47.8 & 61.9 & 66.4 & 65.2 & 74 & 56 & 59.7 \\
    Loose A,B,C & 5 & 65.1 & 73.8 & 74.2 & 49.9 & 35.3 & 49.9 & 70 & 66.1 & 67.6 & 74.9 & 70.4 & 79 \\
    A always before B,C & 5 & 56.8 & 63.5 & 60.5 & 52.4 & 34.4 & 45.7 & 68.7 & 71.4 & 65 & 71.4 & 59.5 & 65 \\
    \midrule
    Mean Acc. & & \textbf{55.7} & 53 & 52.3 & \textbf{41.4} & 33.9 & 36.9 & \textbf{60.7} & 50 & 52.7 & \textbf{68.5} & 57.6 & 52.7 \\
    \bottomrule
    \end{tabu}
    }
    \caption{Zero-Shot Results for Multi-choice TLQA-S over four datasets evaluated on three models, SeViLa~\citep{yu2023sevila} IGVLM~\citep{kim2024igvlm} and LLoVI~\cite{zhang2023llovi}. }
    \label{tab:zs_mcq}
\end{table*}

\subsubsection{Automatic Negative QA Generation}
Given the template questions for a temporal category, along with the positive actions set and instance states, we now present the negative QA generation process. 
To generate negative questions, we take the list of all unique actions for each video and remove the positive actions to get a list of negative actions for that video. These form hard-negative actions set, as the actions are relevant to the video but are negatives. For categories, where  the positive actions list matches all unique actions in video, we take the full action set in the dataset and remove the positive actions set to generate negative actions list.
To generate a balanced split of negative questions for boolean category, we sample an equal number of negatives to match the number of positive questions. To transform the negative action list to questions, we replace the <actor> and <action> in the template question with the computed negative actions, and the answer is set to `no' for boolean question.
For multiple-choice, we randomly sample three negative options from the negative actions list to generate the choices. %Note that the randomly selected negatives can 
Note that we do not add `no' or `not' to generate negative questions, instead we generate questions that are not true for the video as discussed.

% Talk about multiple-choice options - semantic options vs others, guessing easier in the options?

\vspace{-0.2cm}

\section{Dataset Details}
\label{sec:data_details}
\paragraph{Benchamrking variants}

For evaluation, we define 2 versions of the balanced TLQA benchmark to provide multi-scale inference. We define the splits based on number of samples per category in Table~\ref{tab:syntax}: (i) TLQA-Small (TLQA-S) - each category has 2k samples per dataset, %(ii) TLQA-Medium (TLQA-M) - each category has 5k samples per dataset, 
(ii) TLQA-Large (TLQA-L) - each category has $\sim$ 10k samples per dataset. Note that for each split and for each category, we balance the positive and negative questions equally. We provide links to access the data in supplementary.
%For benchmarking, we do zero-shot evaluations with VideoLLaVA~\cite{lin2023videollava}, VideoChatGPT~\cite{Maaz2023VideoChatGPT}, ChatUnivi~\cite{jin2023chatunivi}, SeViLA~\cite{yu2023sevila}, IGVLM~\cite{kim2024igvlm}, LLoVI~\cite{zhang2023llovi}. Note for all LLaVA-based~\cite{liu2023llava} models we use the 7b backbone and SeViLA employs BLIP-2~\cite{li2023blip2} with FlanT5XL.

% \textcolor{red}{move to implementation or related works - add notations for dataset and dealing with backgrounds}
\textbf{STAR}~\cite{wu2021star_situated_reasoning} is a multi-choice VideoQA benchmark for Situated Reasoning with dense scene-graph annotations for object actions and attribute relationships. STAR contains 3k videos consisting of 22K video clips having an average length of 12s.
\textbf{AGQA}~\cite{GrundeMcLaughlin2021AGQA} is an open-ended VideoQA benchmark with compositional spatio-temporal reasoning. It has around 9.7K videos with an average length of 30 seconds with scene-graph annotations. % and temporal questions on before/after/first/last. 
% \textbf{ANetQA}~\cite{yu2023anetqa} is a recent benchmark which has 11.5K untrimmed long videos with fine-grained compositional reasoning questions. The annotations include scene-graph and dense captions which covers actions in the videos. 
\textbf{Breakfast}~\cite{Kuehne12bf} has \textit{fine-grained} temporal annotations for 1.2k cooking videos with average length of 2.3 minutes. \textbf{CrossTask}~\cite{Zhukov2019crosstask} has 2.7k videos with temporal action annotations with an average length of 4 minutes 57 seconds. For both Breakfast and CrossTask dataset we utilize the temporal annotations to generate QA pairs. By employing these diverse datasets, we cover wide range of actions and situations. Note that there are no overlapping actions in Breakfast and CrossTask datasets. Check supplementary Section 1 for more details.

\section{Experiments}
We evaluate the proposed benchmark on various state-of-the-art models.  %For evaluation on our multi-choice TLQA benchmark, we employ the models designed for multiple-choice QA ~\cite{yu2023sevila, kim2024igvlm}.
For benchmarking, we do zero-shot evaluations with VideoLLaVA~\cite{lin2023videollava}, VideoChatGPT~\cite{Maaz2023VideoChatGPT}, ChatUnivi~\cite{jin2023chatunivi}, SeViLA~\cite{yu2023sevila}, IGVLM~\cite{kim2024igvlm}, LLoVI~\cite{zhang2023llovi}. Note for all LLaVA-based~\cite{liu2023llava} models we use the 7b backbone and SeViLA employs BLIP-2~\cite{li2023blip2} with FlanT5XL. We present more evaluations in Supplementary Section 2, we also report performance on LITA~\cite{huang2025lita} and LLaVAOneVision~\cite{li2024llavaOV}. Further, we perform Instruction fine-tuning using TLQA Benchmark and report performance in Supplementary Section 2.

\subsection{Experimental Setup}

\begin{table*} \setlength{\tabcolsep}{5pt}
  \centering
  \resizebox{\textwidth}{!}{
  \begin{tabu}{lccccccccccccc}
    \toprule
    % & & \multicolumn{4}{c}{VQAv2 Accuracy}                   \\
    % \cmidrule(r){3-6}
    Temporal & Complexity &  \multicolumn{3}{c}{STAR}  &  \multicolumn{3}{c}{Breakfast} &  \multicolumn{3}{c}{AGQA} &  \multicolumn{3}{c}{CrossTask}    \\
    % \cmidrule(lr){3-5} \cmidrule(lr){6-8} \cmidrule(lr){9-11}  
    % Operators &  Level &  VideoLLaVA &VideoChatGPT & ChatUnivi      & VideoLLaVA &VideoChatGPT & ChatUnivi       & VideoLLaVA &VideoChatGPT & ChatUnivi           \\
    \cmidrule(lr){3-5} \cmidrule(lr){6-8} \cmidrule(lr){9-11} \cmidrule(lr){12-14}  
    Operators &  Level &  VL &VCG & CV      & VL &VCG & CV       & VL &VCG & CV    & VL &VCG & CV       \\

    % \cmidrule(lr){3-5} \cmidrule(lr){6-8} \cmidrule(lr){9-11} \cmidrule(lr){12-14} \cmidrule(lr){15-17} \cmidrule(lr){18-20} \cmidrule(lr){21-23} \cmidrule(lr){24-26} \cmidrule(lr){27-29}
    
     & & Acc.   & Acc.     & Acc.     & Acc.        & Acc.     & Acc. &    Acc. & Acc.    & Acc. &    Acc. & Acc.    & Acc.  \\
       
        % &  &  &  &  &  &  &   &  &  & & &  & & &   & & &  \\      
    \midrule
    Eventual & 1    & 51.1   & 50.2     &  56.3   & 50.1       & 51.3  &   51.1&     51.5  &  50.2       &  55.2    & 53.6 & 50.7 & 54.2\\
    Always & 2    & 60.7  &  50.7    & 54.5     & -    &  -  &   -  &   61.6 &  46.9  & 52.6 & - & - & -\\
    
    \midrule
    Until & 3    & 52.9   & 50.9    &  59.4   &  50.6     & 51.4   & 54.1  &51.3 & 50.2 &  60.3  & 52.5 &  51.6  & 55.7\\
    Since & 3    & 54.9     & 50.8  & 55.1  & -   & -   & -  &    55.2  & 51.2   & 54.7  & - & - & -\\
    Implies & 3    & 57.3 &  62.2   &  52.6  &    -   & -  &   -  &  59.6      & 61.3      & 53.4  & - & - & -\\
    Before & 3 &  51.5    & 50.5   & 53.3  &  50.3 & 51.2     & 51.6      & 50.9      & 51       & 55.5  & 55.6 &  51.3  & 56.9 \\
    Next & 3 &  52.1   & 51.1   & 56.5   & 51.5  & 52.3  &  54.8   & 51.3  & 50.8     &  55.6  & 53.1  &   51.2  & 54.3\\
    Co-occur & 3    & 51.1     & 53.4     & 58.1    &  -      & -      & -     & 50.2       & 54.7   & 57.6  & - & - & -\\
    Disjoint & 3    & 51.8   & 53.1     & 46.9    & 50.9       & 50.9      & 51.8     & 51.2     & 55.6      &  49.1  & 50.4  &   51.7  & 54\\
    \midrule
    Always Before & 4    & 49  & 49.8   & 49.1   & 47.9    & 48.1  & 50.3   &  49   &50.7    &   50.4  & 48.3   & 49.7  & 51.3  \\
    Always Next & 4    & 49.4  &  49.6  & 49.6   & 47.7  & 48.6   & 48.2   & 50.5    &49.9   &  47.7  & 48.6    & 49.5   & 49.3 \\
    Always Co-occur & 4    & 51.7  & 52.6    & 53.8   &  -  & -   &  - & 56.3        & 52.2  &  56.6  & - & - & -\\
    \midrule
    Strict A,B,C & 5    & 53  & 51.4    & 56.1   & 51.3 &  50.1   & 51.9  & 50.3 & 55     & 56.3   & 51.6  &   56.6  & 51.6\\
    Loose A,B,C & 5 & 51.5   & 50.8   & 52.7   & 51.1  & 49.6 & 50.8 & 50.6   &  50.4     & 55.2  & 53.3 &   49.6  & 52.9 \\
    A always before B,C & 5  & 51.2   & 50.9  & 55.1    & 51.2 &  50.1  &  52.9  & 51.1    &  50.5   & 56.6 & 51.7   &   53.6  & 49.1 \\
    \midrule
    Mean Acc & & 52.6 & 	51.9 & \textbf{	53.9}	 & 50.3	 & 50.4	 & \textbf{51.8} & 	52.7	 & 52	 & \textbf{54.4}  & 51.9 & 51.6 & \textbf{52.9}\\
    \bottomrule
  \end{tabu}
  }
  \caption{Zero-Shot Results on Boolean QA TLQA-S over four datasets. The models are: VL: VideoLLaVA~\cite{lin2023videollava}, VCG: VideoChatGPT~\cite{maaz2023video}, CV: ChatUniVi~\cite{jin2023chat}. It shows that binary QA performance is in general closer to the random base performance than multiple choice QA, indicating a harder task. }
  \label{tab:tlr_small}
  % \vspace{-0.2in}
\end{table*}

% For benchmarking on TLQA, which is composed of boolean and multiple-choice QA, we evaluate on variety of models covering different temporal capacity (using 8/32/100 frames per videos ), general models, specialized models for choice-QA, caption-based approaches. We present more results in supplementary for crosstask dataset and some simple baselines for multiple-choice QA along with caption-based approaches.

For benchmarking on TLQA, which includes both boolean and multiple-choice QA, we evaluate a variety of models with different temporal capacities (using 8/32/100 frames per video), general models, models specialized for multiple-choice QA, and caption-based approaches. Additional results are provided in the supplementary material Section 3, including CrossTask dataset and simple baselines for multiple-choice QA alongside caption-based approaches.

\begin{figure*}
\begin{center}
    \begin{subfigure}[t]{0.48\linewidth}
\includegraphics[width=1\linewidth]{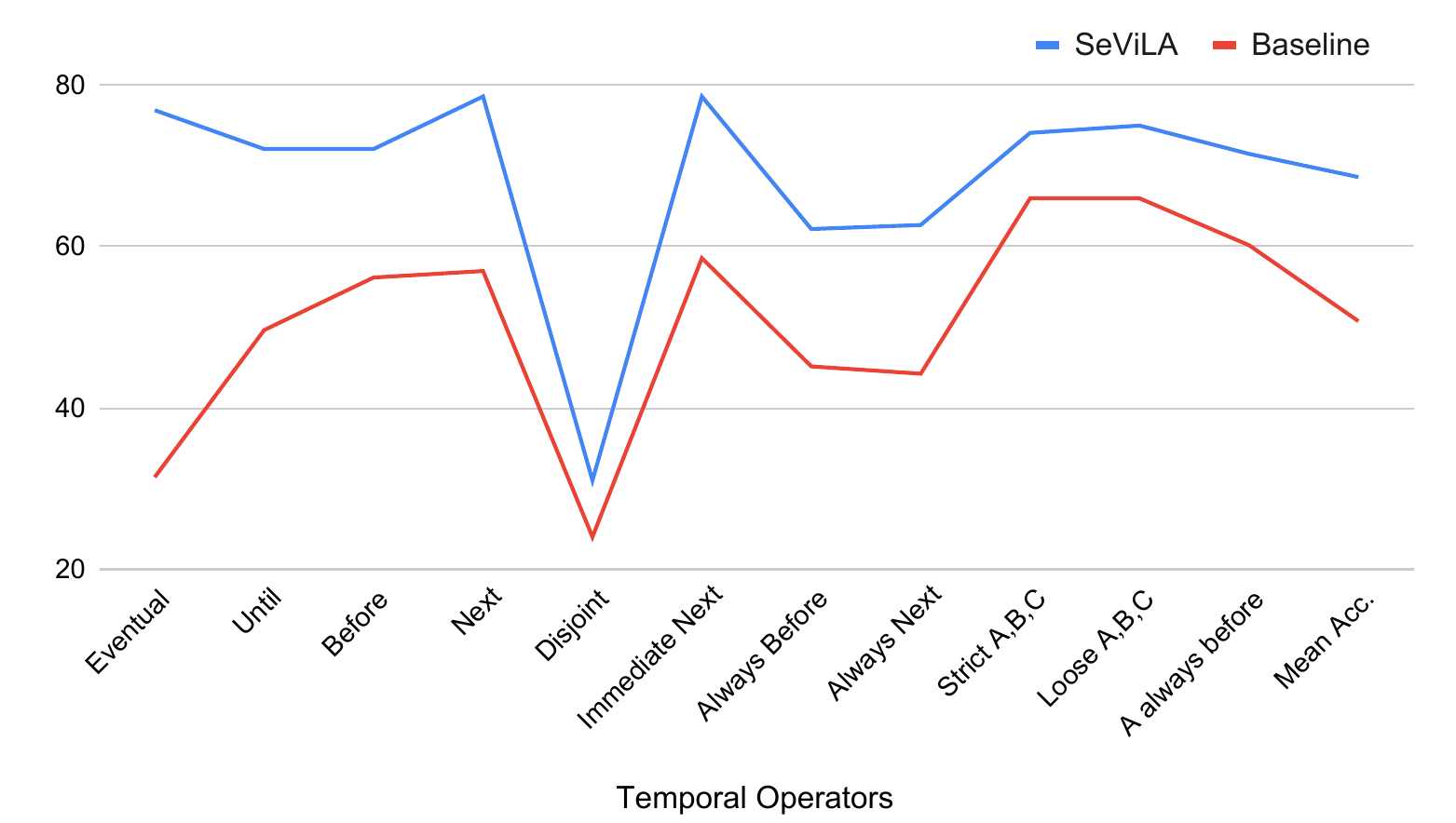}
\caption{MC Baseline for TLQA-S on CrossTask}
\end{subfigure}%
\begin{subfigure}[t]{0.48\linewidth}
\includegraphics[width=1\linewidth]{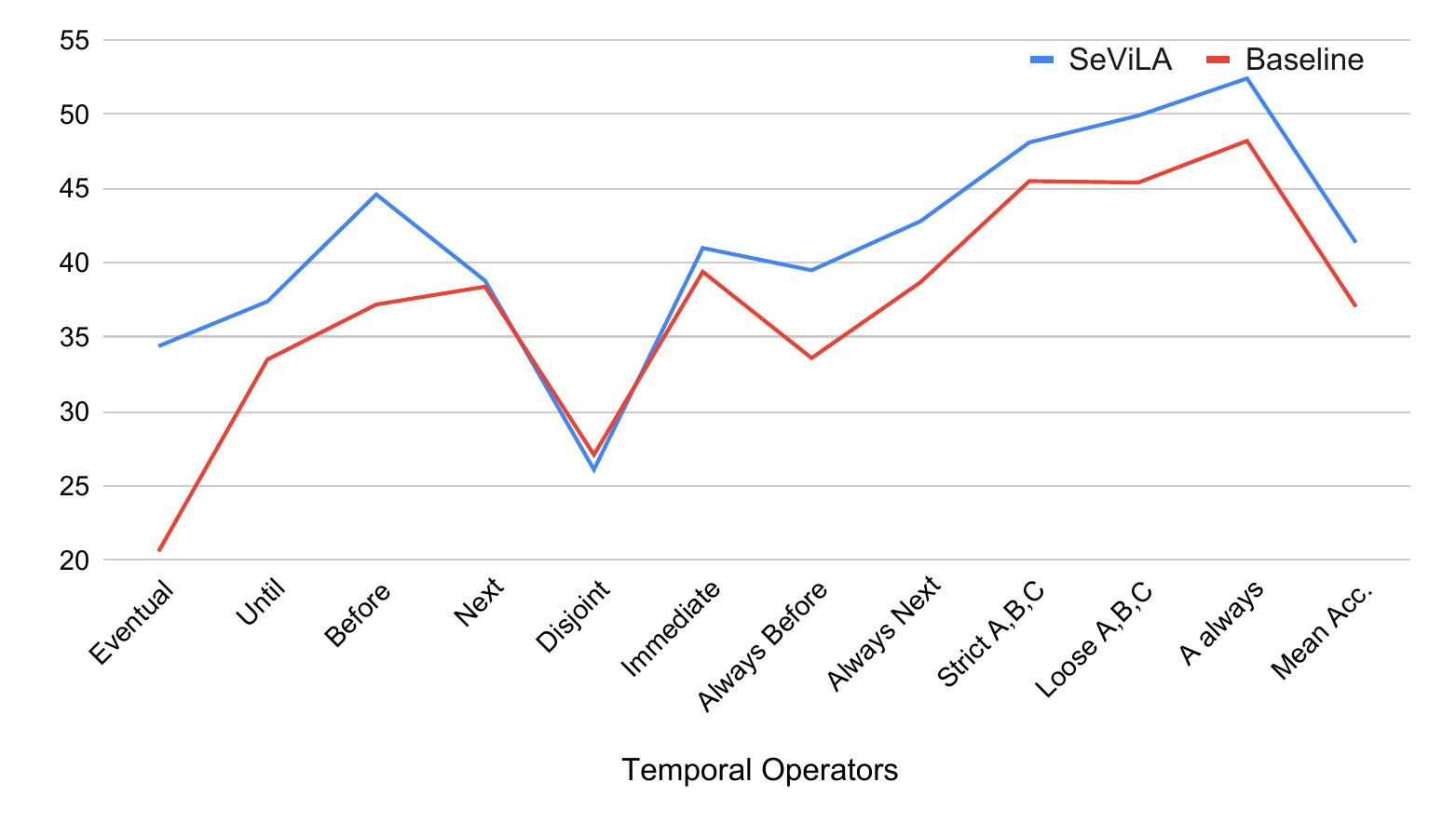}
\caption{MC Baseline for TLQA-S on Breakfast}
\end{subfigure}%
\end{center}
\vspace{-0.1cm}
\caption{Baseline comparison for multiple-choice TLQA. We provide blank frames to SeViLA as a baseline to evaluate the performance on multiple-choice TLQA benchmark. MC: Multiple-Choice.
}
\label{fig:mcq_bs_sevila}
\end{figure*}

For boolean QA benchmark, we perform evaluation on general models like Video-LLaVA~\cite{lin2023videollava}, Video-ChatGPT~\cite{Maaz2023VideoChatGPT} and ChatUniVi~\cite{jin2023chatunivi} that are trained on conversations and caption datasets. We prompt these models to generate yes or no responses by appending the following prompt to the question, `Answer in yes or no only'. We then compute the performance by comparing the model response with the actual answer and report Accuracy. We report F1-score and $\%$Yes ratio which captures the number of times the model responded with a yes answer in the Supplementary Section 3. These metrics will provide insights to overall model performance for boolean QA. Please refer to Supplementary Section 3 for more details.

For more evaluations, please refer to Supplementary section 2 and we perform instruction fine-tuning on TLQA benchmark and report improved performance in Supplementary Section 3. We present more qualitative samples in Supplementary section 7.

% For benchmarking on TLQA boolean QA, we prompt all the models to answer yes or no. We add the following prompt to the question ` Answer in yes or no only.' We perform this benchmarking on Video-LLaVA~\cite{lin2023videollava}, Video-ChatGPT~\cite{Maaz2023VideoChatGPT}, ChatUniVi~\cite{jin2023chatunivi}

% \begin{figure*}
%     \centering
%     \includegraphics[width=1.0\textwidth]{images/q1.png}
%     \caption{Caption}
%     \label{fig:qual}
% \end{figure*}

\begin{figure*}
  \centering
    \begin{subfigure}[]{\textwidth}
        \includegraphics[width=\textwidth]{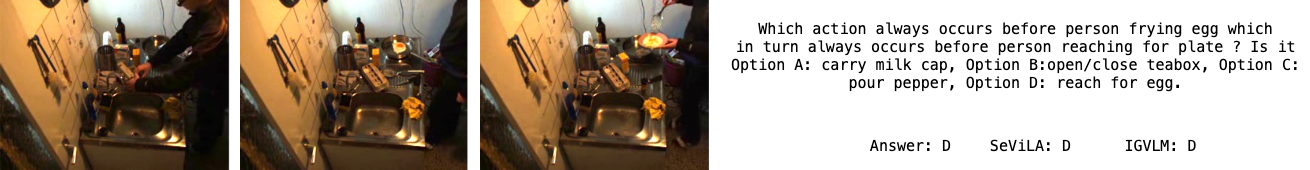}
    \caption{}
    \end{subfigure}
    \begin{subfigure}[]{\textwidth}
        \includegraphics[width=\textwidth]{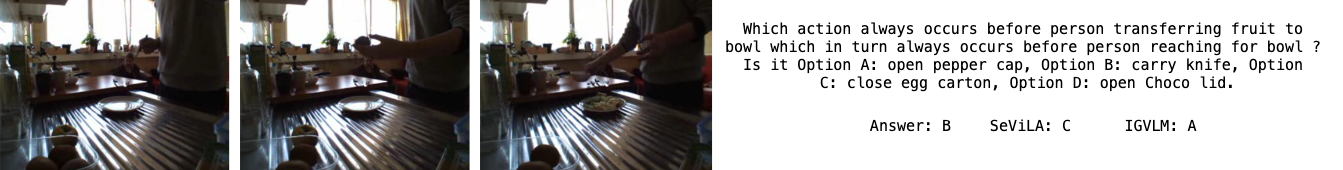}
        \caption{}
    \end{subfigure}
    \begin{subfigure}[]{\textwidth}
        \includegraphics[width=\textwidth]{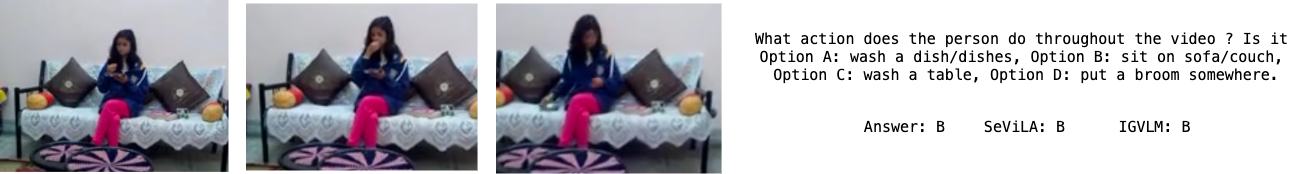}
        \caption{}
    \end{subfigure}

  \caption{Qualitative Results for Multiple-Choice QA: It shows that in some cases, scene or object information might correlate with the correct answer, thus resulting in a easier setup compared to binary QA.  }
  \label{fig:qual}
  %\vspace{-10pt}
\end{figure*}

\vspace{-0.1cm}
\subsection{Results}
We present results for multi-choice TLQA in Table~\ref{tab:zs_mcq}. Overall SeViLA demonstrates superior performance across most temporal operators and datasets compared to IGVLM. Note that SeViLA takes 32 frames as input which is higher temporal capacity than IGVLM. The STAR dataset generally yields higher accuracy for both models, whereas the Breakfast dataset presents more challenges. We present more results in Supplementary Section 3.
In Figure~\ref{fig:qual}, we show some qualitative examples for multiple-choice QA from Breakfast and STAR dataset.

% \begin{wraptable}{r}{0.4\textwidth}
% \captionsetup{font=scriptsize}
% \caption{Zero-Shot Results on Boolean QA TLQA-S for star dataset. }
% \label{tab:star_bool}
% \resizebox{0.4\textwidth}{!}{
% \begin{tabu}{ccccc}
%     \toprule
%     Temporal & Complexity & \multicolumn{3}{c}{} \\
%     \cmidrule(lr){3-5} 
%     Operators & Level & VideoLLaVA & VideoChatGPT & ChatUniVi  \\
%     \midrule
%     Eventual & 1 & 51.1 & 50.2 &     \\
%     Always & 2 & 60.7 & 50.7 &  54.5  \\
%     % \midrule
%     Until & 3 & 52.9 & 50.9 & 59.4    \\
%     Since & 3 & 54.9 & 50.8&  55.1  \\
%     Implies & 3 & 57.3 &62.2 &    \\
%     Before & 3 & 51.5  & 50.5 & 53.3  \\
%     Next & 3 & 52.1 & 51.1 & 56.5     \\
%     Co-Occur & 3 & 51.1 & 53.4 &  58.1    \\
%     Disjoint & 3 & 51.8 & 53.1 &  46.9  \\
%     % % \midrule
%     % Immediate Next & 4 & 49.4 & - & 35   \\
%     % % Always Before & 4 & & &      &  &  &           &   &    &    &   &  &  \\
%     % % Always Next & 4 & & &      &  &  &           &   &    &    &   &  &  \\
%     % Always Co-Occur & 4 &  69.3 & 68.1 & 69.8  \\
%     % % \midrule
%     % Strict A,B,C  & 5 & 48.4 & 60 &  60.5     \\
%     % Loose A,B,C & 5 & 65.1& 73.8 &  74.2     \\
%     % A always before B,C & 5 & 56.8 & 63.5 & 60.5   \\
%     \midrule
%     \textbf{Average} & &   &  & \\
%     \bottomrule
%     \end{tabu}}
% \end{wraptable}

% Results on Choice QA

The table \ref {tab:tlr_small} presents the zero-shot results for Boolean QA TLQA-S across three datasets (STAR, Breakfast, and AGQA) and three models (VideoLLaVA, VideoChatGPT, and ChatUnivi) for various temporal operators. Overall, ChatUnivi performs marginally better than VideoLLaVA and VideoChatGPT. Note that ChatUnivi can process up to 100 frames, whereas both VideoLLaVA and VideoChatGPT are limited to only 8 frames. Despite this, the performance for the boolean category is close to random, highlighting the inherent challenges. This can be attributed to the very high `yes' ratio for VideoLLaVA and VideoChatGPT, while ChatUnivi, despite having a lower 'yes' ratio, also exhibits near-random performance.
We find that the performance trend on TLQA-S and TLQA-L are very similar, we report the details results for TLQA-L in Supplementary Section 6.
Across both boolean and multiple-choice QA, we observe that models with higher temporal capacity tend to perform better. The performance trends on TLQA-S and TLQA-L are very similar. Detailed results for TLQA-L are provided in Supplementary Section 6.

LLoVI~\cite{zhang2023llovi} leverages video captions to perform VideoQA task using LLMs.
For LLoVI~\cite{zhang2023llovi} evaluation on TLQA-S, we extract captions from video at 0.5 fps using LLaVA 7B and employ GPT-3.5-Turbo-1106 model. Following the protocol outlined in ~\cite{zhang2023llovi}, we generate captions and prompt GPT3.5 to evaluate for QA task.
Additionally, we conduct a simple baseline evaluation for Multiple-Choice QA to examine the significance of choices in option selection. For this baseline, we omit the original frames and instead provide blank frames to SeViLA~\cite{yu2023sevila} along with question and choice options. %We report the performance in Table~\ref{tab:supp_zs_mcq_bs}. 
As shown in Figure~\ref{fig:mcq_bs_sevila}, the baseline performance is lower than SeViLA demonstrating that frames aid in better understanding, however note that despite not having any frame information the baseline seems to do well generally highlighting the LLMs ability to pick a choice given 4 choices and question. This is more evident in cases where the question is very generic like `what does the person do in this video?' - the performance of baseline is close to random while SeViLA significantly performs better. For questions that include information about other actions specifically as you go for higher complexity the baseline performance gets better due to LLMs knowledge.

\section{Conclusion}

In this paper, we have introduced the TimeLogic QA (TLQA) framework and benchmark, designed specifically to evaluate the temporal logical understanding capabilities of VideoQA models. 
Our findings underscore the necessity for advancements in modeling temporal understanding in VideoQA. The TLQA benchmark provides a structured and comprehensive testbed, offering a more challenging and realistic evaluation of temporal logical reasoning in video analysis. This framework is scalable and adaptable to various video datasets with temporal annotations, enabling the transformation of any annotated video dataset into temporal logic QA pairs.
% We hope that the proposed framework will help to advance temporal logical understanding in video by providing a challenging and comprehensive evaluation standard.
It is our contention that the future of VQA, and by extension video comprehension, lies in the development of models that can navigate the temporal dimension with the same agility that humans do. 
{
    \small
    \bibliographystyle{ieeenat_fullname}
    \bibliography{main}
}


\end{document}